\documentclass[11pt]{article}
\usepackage{latex/acl}

\newcommand{\modelname}{\texttt{LORex}}

\newcommand{\discri}{\texttt{TEMPORA}}

\usepackage{caption}
\usepackage[linesnumbered,ruled,vlined]{algorithm2e}
\usepackage{orcidlink}
\usepackage{times}
\usepackage{latexsym}
\usepackage{hyperref}   

\usepackage{fdsymbol}

\usepackage{amssymb}
\usepackage{graphicx} 
\usepackage{todonotes}
\usepackage{multirow}
\usepackage{multicol}
\usepackage{tcolorbox}

\usepackage[T1]{fontenc}

\usepackage[utf8]{inputenc}
\usepackage{booktabs}
\usepackage{graphicx}
\usepackage{multirow}
\usepackage{microtype}
\usepackage{makecell}
\usepackage{longtable}
\usepackage{enumitem}

\usepackage{inconsolata}

\usepackage{enumitem}
\newtheorem{definition}{Definition}

\newif\ifupdate\updatetrue

\usepackage{graphicx}
\usepackage{amsmath}
\usepackage{subcaption}

\title{\textit{Rank, Chunk, and Expand}: \\Lineage-Oriented Reasoning for Taxonomy Expansion}

\author{
  Sahil Mishra \orcidlink{0000-0001-5477-9003}\textsuperscript{\scalebox{1.5}{$\clubsuit$}}\thanks{\ \ Corresponding author.} \hspace{2.5em}
  Kumar Arjun \orcidlink{0009-0005-4912-5757}\textsuperscript{\scalebox{1.5}{$\diamondsuit$}} \hspace{2.5em} 
  Tanmoy Chakraborty \orcidlink{0000-0002-0210-0369}\textsuperscript{\scalebox{1.5}{$\clubsuit$}} \\
  \textsuperscript{\scalebox{1.5}{$\clubsuit$}}Department of Electrical Engineering, IIT Delhi \space
  \textsuperscript{\scalebox{1.5}{$\diamondsuit$}}Department of Mathematics, IIT Delhi \\
  \textsuperscript{\scalebox{1.5}{$\clubsuit$}}\texttt{\{sahil.mishra, tanchak\}@ee.iitd.ac.in, \textsuperscript{\scalebox{1.5}{$\diamondsuit$}}kumar.arjun.mt121@maths.iitd.ac.in}
}

\newtcolorbox[auto counter]{mybox1}[2][]{%
    colback=blue!5!white, colframe=blue!75!black, fonttitle=\bfseries, title=#2, #1
}

\newtcolorbox[auto counter]{mybox2}[2][]{%
    colback=blue!5!white, colframe=blue!75!black, fonttitle=\bfseries, title=#2, #1
}

\newtcolorbox[auto counter]{promptbox}[2][]{
    colback=blue!5!white, colframe=blue!75!black, 
    fonttitle=\bfseries, title=#2, #1, label=#2, , width=1.0\textwidth
}

\begin{document}
\maketitle
\begin{abstract}

{Taxonomies are hierarchical knowledge graphs crucial for search engines, recommendation systems, and web applications. As data grows, expanding taxonomies is essential, but existing methods face key challenges: (1) discriminative models struggle with representation limits and generalization, while (2) generative methods either process all candidates at once, introducing noise and exceeding context limits, or discard relevant entities by selecting a sample of candidates. We propose \modelname\ (\textbf{L}ineage-\textbf{O}riented \textbf{Re}asoning for Taxonomy E\textbf{x}pansion), a plug-and-play framework that combines discriminative ranking and generative reasoning for efficient taxonomy expansion without fine-tuning. Unlike prior methods, \modelname\ ranks and chunks candidate terms into batches, filtering noise and iteratively refining selections by reasoning candidates' hierarchy to ensure contextual efficiency. Extensive experiments across four benchmarks and twelve baselines show that \modelname\ improves accuracy by 12\% and Wu \& Palmer similarity 5\% over state-of-the-art methods.}
\end{abstract}

\section{Introduction}
Taxonomies are hierarchical graph structures that capture hypernymy (``\textit{is-a}'') relationships among concepts, making them essential for knowledge organization across various domains. Conglomerates leverage them to power search engines \cite{strzelecki2019snippets,janssen2021functionality}, enhance product recommendations \cite{mao2020octet, karamanolakis_txtract_2020} and improve advertisements \cite{owl, manzoor2020expanding}. Despite their utility, real-world taxonomies are manually created by domain experts, which is cumbersome and costly, limiting their ability to capture emerging concepts \cite{jurgens2016semeval, bordea2016semeval}. Additionally, as new concepts continuously emerge, updating taxonomies manually becomes increasingly impractical.

\begin{figure}[!ht]
\centering
\includegraphics[width=1\linewidth]{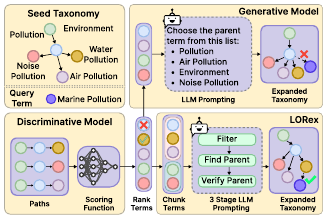}
\caption{Illustration of taxonomy expansion task and contribution of \modelname\ framework.}
\label{fig:intro}
\end{figure}

To overcome the limitations of manual taxonomy construction, research has now shifted toward taxonomy expansion, which involves the integration of new entities into an existing seed taxonomy \cite{jurgens2016semeval,bordea2016semeval}. This process positions new concepts under appropriate existing nodes, known as \textit{anchor nodes}, to preserve the structural integrity of the original taxonomy. As illustrated in Fig. \ref{fig:intro}, this ensures that new terms, like ``Marine Pollution'', are inserted under ``Water Pollution'' rather than the broader ``Pollution'', maintaining hierarchical coherence.

Research on taxonomy expansion has evolved through three distinct generations. The first-generation approaches primarily rely on the seed taxonomy as weak supervision to assess hypernymy relationships, leveraging lexical pattern matching \cite{jurgens-pilehvar-2015-reserating} and distributional embeddings \cite{chang2017distributional}. However, these methods are inherently constrained by limited self-supervised annotation data and, therefore, fail to fully harness the hierarchical and structural intricacies embedded within taxonomies.

Second-generation research improves taxonomy expansion by incorporating structural summaries like mini-paths \cite{yu_steam_2020, jiang2022taxoenrich}, paths \cite{liu2021temp}, random walks \cite{xutaxoprompt}, ego-nets \cite{shen2020taxoexpan}, and local graphs \cite{wang2021enquire} to enrich hierarchical representation. These methods are also supplemented with external knowledge, such as domain-specific corpora \cite{yu_steam_2020} and concept definitions \cite{liu2021temp}. However, they struggle with small-scale taxonomies, where limited self-supervised data hinders model training and generalization, requiring more scalable approaches.

The current generation research on taxonomy expansion leverages LLMs to integrate new concepts into existing taxonomies. Models like LLaMA \cite{touvron2023llama, dubey2024llama} and GPT-4 \cite{achiam2023gpt}, excel in structural knowledge comprehension \cite{kicgpt, liu2025filter}. Notably, GPT-4, with 1.7 trillion parameters, generates parent terms directly \cite{zeng2024codetaxo, zeng2024chain}, while LLaMA-2 (7B parameters) requires fine-tuning on self-supervised annotation data for taxonomy expansion \cite{flame2025, xu2025compress}.

While effective, these approaches incur high computational costs. Fine-tuning LLaMA models requires substantial resources, whereas GPT-4 inference is expensive due to its massive parameter size and paid API-based access, further increasing costs and reducing flexibility for scaling taxonomy expansion. Additionally, LLM-driven methods face context size limitations -- some attempt to encode all candidates into prompts \cite{flame2025}, which is impractical, while others rely on top-k selection \cite{zeng2024codetaxo}, risking the omission of relevant terms.

To address limitations of existing taxonomy expansion methods, we propose \textbf{\modelname} (\textbf{L}ineage-\textbf{O}riented \textbf{Re}asoning for Taxonomy E\textbf{x}pansion), a plug-and-play that integrates discriminative ranking and generative reasoning to systematically expand taxonomies without fine-tuning LLMs or discarding relevant candidates.

Inspired by \citet{liu2021temp}, \modelname\ first ranks the candidate terms for an entity using a discriminative ranker, named \discri, which leverages lineage structures to score and rank candidates. \discri\  converts the lineage into a taxonomy path via an Euler tour, then verbalizes relationships by labeling edges as \textit{``parent of''} and \textit{``child of''}. Ranked candidates are then chunked into batches, where an LLM first filters relevant batches and performs hierarchical reasoning to select the best parent. To verify the correctness, we introduce an LLM-based path-scoring function that ranks candidates based on their lineage structure.

We conduct extensive experiments to evaluate \discri\ for candidate ranking and \modelname\ for taxonomy expansion on four public benchmark datasets, including SemEval-2016 Task 13 Taxonomy Extraction Evaluation \cite{bordea2016semeval} and WordNet \cite{bansal-etal-2014-structured}. \discri\ outperforms all ranking baselines, improving Hit@k by 21.3\%, despite their extensive self-supervised training. \modelname\ surpasses eight baselines across four benchmarks, achieving 12\% higher accuracy and a 5\% gain in Wu \& Palmer (Wu\&P) metric. Furthermore, we perform ablations to analyze the performance of \modelname\ under varying conditions, including its effectiveness with different open-source LLMs.

We summarize our main contributions below:\footnote{Code: https://github.com/sahilmishra0012/LOREx.}
\begin{itemize}[noitemsep, nolistsep, topsep=0pt, leftmargin=1em]
\setlength{\itemsep}{0pt}
    \item We introduce \modelname, a plug-and-play framework that combines discriminative and generative methods to rank and chunk candidates, followed by iterative reasoning over hierarchy to retrieve and verify parent terms for taxonomy expansion without fine-tuning LLMs or discarding relevant candidates.
    \item We develop \discri, a discriminative ranker, which transforms lineage into path by performing an Euler tour on it, followed by path verbalization, where taxonomy edges are annotated with relational labels such as ``parent of'' and ``child of'' to enhance interpretability.
    \item Experiments on four benchmark datasets demonstrate that \discri\ surpasses all ranking baselines with a 21.3\% improvement in Hit@k, while \modelname\ achieves a 12\% increase in accuracy and a 5\% boost in Wu\&P metric, establishing an efficient state-of-the-art in taxonomy expansion.
\end{itemize}

\begin{figure*}[!ht]
  \centering
\includegraphics[width=1.0\linewidth]{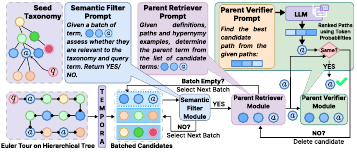}
  \caption{An illustration of our proposed framework, \modelname, which first ranks candidates using \discri\ and chunks them. The semantic filter module then checks the batch relevance, which further leads to parent retrieval and verification. In Euler tour, \(\bullet\) (\(\blacklozenge\)) means ``\textit{is child of}'' (``\textit{is parent of}'') relationship.}
  \label{fig:model}
\end{figure*}

\section{Related Work}
\textbf{Taxonomy Expansion.} Expert-curated taxonomies, such as The Plant List \cite{TPL} and NCBI Taxonomy \cite{NCBI}, require continuous updates, necessitating robust taxonomy expansion. Recent discriminative methods leverage structural features such as taxonomy paths \cite{liu2021temp}, mini-paths \cite{yu_steam_2020}, hierarchical neighbourhood\cite{jiang2022taxoenrich}, ego-trees \cite{wang2021enquire}, ego networks \cite{shen2020taxoexpan} and even quantum entanglements \cite{mishra2025quantaxo}, to enhance anchor node representation and model hypernymy relationships. Prompting frameworks such as TaxoPrompt \cite{xutaxoprompt}, ATTEMPT \cite{2023attempt}, and TacoPrompt \cite{xu-etal-2023-tacoprompt} model taxonomy paths via masked language modeling. However, these methods struggle with limited self-supervised annotated data and small-scale encoders. To address this, LLM-based solutions such as FLAME \cite{flame2025} and COMI \cite{xu2025compress} fine-tune LLaMA models, while CodeTaxo \cite{zeng2024codetaxo} employs GPT-4 inference. Though effective, fine-tuning is computationally intensive, and GPT-4 API usage is costly. To address these challenges, our proposed \modelname\ model integrates discriminative ranking with generative reasoning, leveraging smaller language models. \modelname\ employs iterative prompting to systematically reason over taxonomic lineages, ensuring computational efficiency while maintaining high accuracy in taxonomy expansion.

\noindent \textbf{Structure Reasoning using LLMs.} Recent studies on structure reasoning with LLMs have explored graph-based integration to enhance knowledge grounding and reasoning capabilities. Studies such as Think-on-Graph \cite{sun2024thinkongraph} employs iterative beam searches over knowledge graphs (KGs) to retrieve reasoning paths, while Graph-CoT \cite{jin2024graph} extends Chain-of-Thought (CoT) prompting by incorporating graph structures into iterative reasoning cycles. Similarly, Paths-over-Graph \cite{tan2024paths} enhances reasoning over structure by integrating dynamic multi-hop path exploration and pruning techniques within KGs. Additionally, code language prompting methods like CodeKGC \cite{bi2024codekgc} and Code4struct \cite{wang2022code4struct} have also been used to reason over graphs by converting graph entities into classes following object-oriented programming paradigm. Building on these advances, our work focuses on iterative reasoning to enhance structural representation learning.

\section{Problem Definition}
\begin{definition}
    \textbf{Taxonomy:} A taxonomy \(\mathcal{T}^o = (\mathcal{N}^o, \mathcal{E}^o)\) is a hierarchically structured directed acyclic graph (DAG), where each node \(n \in \mathcal{N}^o\) denotes a distinct entity encapsulated by a concept and supplemented with a description \(D_n \in D\), while each edge \(\langle n_p, n_c \rangle \in \mathcal{E}^o\) signifies a ``parent-child'' dependency.
\end{definition}
As new entities continue to emerge, the key challenge is incorporating them into an existing taxonomy \(\mathcal{T}^o\), referred to as taxonomy expansion,
\begin{definition}
    \textbf{Taxonomy Expansion:} Taxonomy expansion is the process of incorporating emergent concepts \(C\) into a seed taxonomy \(\mathcal{T}^o=(\mathcal{N}^o,\mathcal{E}^o)\) to create an updated taxonomy \(\mathcal{T}=(\mathcal{N}^o \cup \mathcal{C},\mathcal{E}^o \cup \mathcal{R})\), where \(\mathcal{R}\) is the set of new relations connecting existing entities \(\mathcal{E}^o\) with the new concepts \(C\).
\end{definition}

\section{Proposed Method}
\label{method}
\modelname, a plug-and-play taxonomy expansion framework (Fig. \ref{fig:model}), integrates discriminative ranking (\discri) with iterative structure reasoning to integrate new concepts into an existing taxonomy. 

\subsection{\discri}
\label{subsec:tempora}
The first stage of \modelname\ is to rank and chunk all candidate nodes \(\mathcal{N}^o\) for all query nodes \(\mathcal{C}\). Since LLMs have context length limitations, incorporating all candidate terms \(\mathcal{N}^o\) in the prompt is impractical and expands the search space. Therefore, we need to rank candidates to prioritize relevance. We propose \discri, a discriminative ranker (inspired by TEMP \cite{liu2021temp}), which uses pre-trained encoders to learn path-based features. Unlike TEMP, \discri: (i) incorporates siblings and children to better capture hierarchical structures, and (ii) verbalizes paths for enhanced interpretability, whereas TEMP relies on special tokens like \texttt{[SEP]} or \texttt{[UNK]}.

To effectively capture the hierarchical structure, we extract the path from the root node to the anchor node, along with its children and siblings. This hierarchy is then verbalized by performing an Euler tour on the tree, which begins at the anchor node, ascends to the root, and subsequently traverses siblings, followed by child nodes in a structured manner. The Euler-toured path is represented as \(P=[n_p, n_{p-1}, \allowbreak \dots, n_o, n_1, \dots, \allowbreak n_{p-1}, n_{s_1},\dots, n_{p-1}, n_p, n_{c_1}, n_p, n_{c_2}]\), where \(n_o\) is the root node while \(n_{p-i}\), \(n_{s_i}\) and \(n_{c_i}\) are the ancestor siblings and children of anchor node, respectively. Unlike traditional Euler tour, where each node can only be visited once, our version of Euler tour can visit a node more than once. Contrary to TEMP, we use interpretable relational phrases such as \textit{``is parent of''} and \textit{``is child of''} to enhance the clarity and interpretability of the Euler-toured path, returning verbalized path as \(P_v\). We return two different verbalized paths $P_{v}(\!q,\!P)\!=\!\textit{Verbalizer}(\!q,\!P)$ and $P_{\!v\!}(P)\!=\!\textit{Verbalizer}(\!P)$ where \(q\) is the query node whose definition is used by the verbalizer, which is concatenated with path \(P\) via the \texttt([SEP]) token. The pre-trained encoder returns the following vectors,
\begin{equation}
    \begin{aligned}
        f(P_v(q,P))\!&=\!v_{[\texttt{CLS}]},\!v_1,\!\ldots,\!v_{[\texttt{SEP}]},\\&\!v_{n_p},\!\ldots,\!v_{{n_{c_2}}}\!,
    \end{aligned}
\end{equation}
\begin{equation}
    f(P_v(P)) = v_{[\texttt{CLS}]}v_{n_p}, \dots, v_{{n_{c_2}}},
\end{equation}
where \(v_{[\texttt{CLS}]}\) represents the vectorized embedding of \texttt{[CLS]} token, which is processed through a Multi-Layer Perceptron (MLP) to compute the fitting score. TEMP is optimized using a dynamic margin loss that differentiates between positive (\(P^+_{v}(q, P)\)) and negative (\(P^-_{v}(q, P)\)) paths. However, TEMP undergoes extensive training across multiple benchmarks, as its primary objective is to predict the most suitable candidate term. In contrast, \discri\ is a simple retriever, which should not need exhaustive training. A minimal number of training epochs should be sufficient for its intended function.

However, limiting training to a few epochs introduces several challenges, as the model struggles to distinguish between positive and negative paths effectively. To mitigate this, we introduce a dual-path training strategy wherein the model is simultaneously trained on \(P_v(q,P)\) and \(P_v(P)\). Specifically, for positive samples, we minimize the margin loss between these paths, while for negative samples, we maximize it. The losses are defined as,
\begin{equation}
    \mathcal{L_m}\!=\!
    \!\sum_{\substack{P,P'}} 
    \!\max(\!0,\!-f(\!P)\!+\!f(\!P')\!+\!\gamma(\!P,\!P')\!)
\end{equation}
\begin{equation}
    \mathcal{L}_{+} = \sum_{\substack{P,P_r}}\big\| f(P) - f(P_r) \big\|^2
\end{equation}
\begin{equation}
    \mathcal{L}_{-} = \sum_{\substack{P',P'_r}}\max \big( 0, f(P') - f(P'_r) \big),
\end{equation}
where \(P \in P_v(q,P)^+\) and \(P' \in P_v(q,P)^-\) are positive and negative verbalized paths with query definitions while \(P_r \in P_v(P)^+\) and \(P'_r \in P_v(P)^-\) exclude definitions.  Visualization of different types of paths is shown in Appendix \ref{app:hierarchy}. The joint loss is represented as,
\begin{equation}
\mathcal{L} = \mathcal{L_m} + \lambda_1 \cdot \mathcal{L}_{+} + \lambda_2 \cdot \mathcal{L}_{-},
\end{equation}
where \(\lambda_1\) and \(\lambda_2\) are hyperparameters controlling the importance of the path difference losses. To measure the semantics of taxonomy paths, we use the dynamic margin function proposed by TEMP,
\begin{equation}
    \gamma(P, P') = \left( \frac{|P \cup P'|}{|P \cap P'|} - 1 \right) \cdot d,
\end{equation}
where \(d\) is the margin adjustment parameter.
Self-supervised data sampling and training is done in the same way as for TEMP. During ranking, we rank all candidate terms \(n_p\! \in\!\mathcal{N}^o\) against all query terms \(n_c\! \in\! \mathcal{C}\) by computing fitting score \(f(P_v(n_q,P_p))\), where \(P_p\) is the Euler-toured path of candidate term \(n_p\). Then, we chunk ranked candidate terms in batches of \(k\). For a query term \(n^i_c\), the batches are \(\mathcal{B}^i\!=\!\{ B^i_1, B^i_2, \dots, B^i_m \}\), where each batch  \(B^i_j\)  contains  \(k\)  elements (except possibly the last batch).

\subsection{Semantic Filter Module}
\label{subsec:semfilt}
After obtaining a batch of candidate terms \(B^i_j\) for the query term \(n^i_c\), it is crucial to assess their relevance to the taxonomy or the query term. The filtering step ensures that only contextually relevant batches are passed for further task, enhancing retrieval precision while reducing computational overhead. For instance, a batch containing terms such as [``environmental policy'', ``environmental degradation'', ``environmental protection''] exhibits strong semantic alignment with the ``environment'' taxonomy, particularly in relation to ``quality of the environment.'' In contrast, a batch comprising [``bird'', ``ornithology'', ``genetics''] lacks the necessary contextual relevance, making it less suitable for parent retrieval.

To achieve this, we introduce a semantic filter module, which ensures that only semantically coherent candidate terms advance to subsequent retrieval and verification stages, effectively reducing processing overhead by discarding irrelevant batches. The filtering process is executed via Boolean inference, where the LLM evaluates the semantic alignment of a batch with the taxonomy or query term. The LLM is prompted with query term \(n^i_c\) along with candidate terms \(B^i_j\) and their definitions \(D^i_j\) to achieve fair semantic evaluation. If the LLM returns `Yes,' the batch proceeds to subsequent processing; otherwise, it is discarded (see Appendix \ref{app:semprompt} for the prompt).

\subsection{Parent Retriever Module}
\label{subsec:parentretriever}
Once the candidate batch \(B^i_j\) successfully passes the semantic filtering step, we utilize the lineage of the candidate terms to determine the most suitable parent for the given query term \(n^i_c\). This process involves reasoning over the hierarchy of the candidate term, as discussed in Section \ref{subsec:tempora}, which involves leveraging path to the root node, ego network and definition of the candidate, to infer the most granular candidate that directly subsumes the query term. The LLM is prompted to select the most suitable hypernym from the candidate list, returning \texttt{NOT FOUND} if no valid hypernym exists (see Appendix \ref{app:parprompt} for the prompt).

However, a key limitation of the retriever is that the LLM does not explicitly return \texttt{NOT FOUND} when no suitable hypernym exists. This occurs because the candidate terms are pre-ranked, leading to semantically similar terms being grouped within the same batch. As a result, the LLM often selects a term that is closely related but not the most appropriate hypernym, failing to discard incorrect candidates and move to the next batch for a better match. To address this, a verification step is needed to ensure the predicted hypernym aligns with the taxonomy structure.

\subsection{Parent Verifier Module}
\label{subsec:parentverifier}

In order to verify if the retrieved parent from LLM is the most appropriate one, existing works have employed discriminative verifiers \cite{zeng2024chain}. However, these verifiers are often limited by their representation power. Moreover, if we use path-based verifier similar to \discri, the verifier is again biased towards the ranked candidates, creating a bottleneck in verification. To overcome this, we instead utilize LLMs to reason over candidate paths within a batch to identify the most plausible hierarchical structure. However, direct reasoning over paths introduces instability, particularly when using smaller instruction-tuned LLMs. These models frequently generate inconsistent or malformed outputs, omitting full paths or introducing noise due to complex reasoning steps. To mitigate this, we compute the average token log-probability for each candidate path in the prompt and use these scores to rank them. This approach offers a more stable and interpretable signal for selecting the best-reasoned path.

The LLM is prompted with the hierarchical path of retrieved candidate from parent retriever module along with the path of remaining candidates within the batch to select the most appropriate path (see Appendix \ref{app:verprompt} for the prompt). Then, average token log probabilities are computed for all the paths in the batch to rank them to eliminate the possibility of the verifier selecting longer paths. The highest-scored candidate path is computed as,
\begin{equation}
P^*_j\!=\!\arg\max_{P_i\in B^i_j}\frac{1}{n}\sum_{i=1}^{n} \log p(t_i | t_1, \dots, t_{i-1}),
\end{equation}
where \(P^*_j\) represents the selected path with the highest cumulative log probability score while \(t_i\) denotes the \(i\)-th token in the path.

If the retrieved candidate path aligns with the verifier's selection, it is retained as the most appropriate hypernym; otherwise, the candidate is removed from the batch, and the parent retriever is re-invoked to select a new hypernym. This iterative process continues until only two candidates remain in the batch. If the verifier fails to correctly identify the retrieved path among the final two candidates, the entire batch is discarded, and the process moves to the next batch. The pseudocode of the framework is provided in Algorithm \ref{algorithm:1} in Appendix \ref{app:pseudocode}.

\section{Experiments}
\renewcommand{\arraystretch}{1.0}
\setlength{\tabcolsep}{3pt}
\begin{table}[!t]
\centering
\begin{tabular}{lcccc}
\toprule
\textbf{Method} & \multicolumn{2}{c}{\textbf{Environment}} & \multicolumn{2}{c}{\textbf{Science}} \\
\textbf{}       & \textbf{Hit@1}     & \textbf{Hit@10}     & \textbf{Hit@1}   & \textbf{Hit@10}   \\
\midrule
BM25            & 0.077              & 0.250               & 0.118            & 0.212             \\
SBERT$_{COS}$   & 0.173              & 0.615               & 0.294            & 0.576             \\
BERT$_{COS}$    & 0.115              & 0.385               & 0.106            & 0.329             \\
SimCSE          & 0.120              & 0.450               & 0.177            & 0.468             \\
TEMP            & \underline{0.403}  & \underline{0.654}   & \underline{0.459} & \underline{0.612} \\ 
\midrule
\discri\        & \textbf{0.481}     & \textbf{0.731}      & \textbf{0.529}   & \textbf{0.753}    \\ 
\bottomrule
\end{tabular}
\caption{Performance Comparison of \discri\ with discriminative ranking baselines. The state of the art is \textbf{bolded}, while the best baseline is \underline{underlined}.}
\label{table:results1}
\end{table}

\renewcommand{\arraystretch}{1.1}
\setlength{\tabcolsep}{5pt}
\begin{table*}[ht]
\centering
\scalebox{0.87}{\begin{tabular}{lcccccccc}
\toprule
\multicolumn{1}{c}{\multirow{2}{*}{\textbf{Methods}}} & \multicolumn{2}{c}{\textbf{SemEval16-Env}} & \multicolumn{2}{c}{\textbf{SemEval16-Sci}} & \multicolumn{2}{c}{\textbf{SemEval16-Food}} & \multicolumn{2}{c}{\textbf{WordNet}} \\
\multicolumn{1}{c}{} & \textbf{Acc} & \textbf{Wu\&P} & \textbf{Acc} & \textbf{Wu\&P} & \textbf{Acc} & \textbf{Wu\&P} & \textbf{Acc} & \textbf{Wu\&P} \\ 
\midrule
BERT+MLP & 12.6 $ \pm $ 1.1 & 48.3 $ \pm $ 0.8 & 12.2 $ \pm $ 1.7 & 45.1 $ \pm $ 1.1 & 12.7 $ \pm $ 1.8 & 49.1 $ \pm $ 1.2 & 9.2 $ \pm $ 1.2 & 43.5 $ \pm $ 0.4 \\
TAXI & 18.5 $ \pm $ 1.3 & 47.7 $ \pm $ 0.4 & 13.8 $ \pm $ 1.4 & 33.1 $ \pm $ 0.7 & 20.9 $ \pm $ 1.1 & 41.6 $ \pm $ 0.2 & 11.5 $ \pm $ 1.8 & 38.7 $ \pm $ 0.7 \\
Musubu & 42.3 $ \pm $ 3.2 & 64.4 $ \pm $ 0.7 & 44.5 $ \pm $ 2.3 & 75.2 $ \pm $ 1.2 & 38.6 $ \pm $ 2.7 & 63.4 $ \pm $ 0.4 & 25.6 $ \pm $ 4.7 & 61.4 $ \pm $ 1.6 \\
TaxoExpan & 10.7 $ \pm $ 4.1 & 48.5 $ \pm $ 1.7 & 24.2 $ \pm $ 5.4 & 55.6 $ \pm $ 1.9 & 24.6 $ \pm $ 4.7 & 52.6 $ \pm $ 2.2 & 17.3 $ \pm $ 3.5 & 57.6 $ \pm $ 1.8 \\
STEAM & 34.1 $ \pm $ 3.4 & 65.2 $ \pm $ 1.4 & 34.8 $ \pm $ 4.5 & 72.1 $ \pm $ 1.7 & 31.8 $ \pm $ 4.3 & 64.8 $ \pm $ 1.2 & 21.4 $ \pm $ 2.8 & 59.8 $ \pm $ 1.3 \\
TEMP & 45.5 $ \pm $ 8.6 & 77.3 $ \pm $ 2.8 & 43.5 $ \pm $ 7.8 & 76.3 $ \pm $ 1.5 & 44.5 $ \pm $ 0.3 & 77.2 $ \pm $ 1.4 & 24.6 $ \pm $ 5.1 & 61.2 $ \pm $ 2.3 \\
HEF & 51.4 $ \pm $ 2.8 & 71.4 $ \pm $ 2.3 & 48.6 $ \pm $ 5.3 & 72.8 $ \pm $ 1.8 & 46.1 $ \pm $ 4.3 & 73.5 $ \pm $ 3.2 & 28.1 $ \pm $ 4.4 & 64.5 $ \pm $ 1.8 \\
BoxTaxo & 32.3 $ \pm $ 5.8 & 73.1 $ \pm $ 1.2 & 26.3 $ \pm $ 4.5 & 61.6 $ \pm $ 1.4 & 28.3 $ \pm $ 5.1 & 64.7 $ \pm $ 1.6 & 22.3 $ \pm $ 3.1 & 58.7 $ \pm $ 1.2 \\
TaxoComplete & 45.3 $ \pm $ 1.7 & 63.4 $ \pm $ 0.2 & 38.4 $ \pm $ 0.7 & 54.8 $ \pm $ 0.3 & 33.6 $ \pm $ 2.5 & 56.1 $ \pm $ 0.6 & 28.3 $ \pm $ 3.9 & 56.2 $ \pm $ 2.1 \\
TacoPrompt & 56.2 $ \pm $ 3.2 & 82.1 $ \pm $ 0.4 & 53.1 $ \pm $ 6.6 & 76.3 $ \pm $ 0.8 & 53.4 $ \pm $ 3.7 & \underline{80.6 $ \pm $ 0.4} & 44.8 $ \pm $ 4.1 & 72.3 $ \pm $ 1.6 \\
TaxoPrompt & 51.9 $ \pm $ 6.3 & 78.6 $ \pm $ 1.7 & 58.3 $ \pm $ 3.8 & 78.1 $ \pm $ 0.7 & 49.5 $ \pm $ 3.7 & 74.4 $ \pm $ 1.4 & 38.5 $ \pm $ 4.5 & 68.2 $ \pm $ 1.2 \\
FLAME & 63.4 $ \pm $ 1.9 & \textbf{85.1 $ \pm $ 0.3} & 63.2 $ \pm $ 4.1 &  \underline{82.5 $ \pm $ 1.2} & \textbf{58.7 $ \pm $ 3.5} & 78.1 $ \pm $ 1.5 & 45.2 $ \pm $ 1.3 & 71.5 $ \pm $ 0.6 \\ 
\midrule
\modelname\(_{7B-2C}\) &  59.7 $ \pm $ 1.3 &  78.5 $ \pm $ 0.5 & 55.3 $ \pm $ 2.5 & 76.4 $ \pm $ 0.2 & 49.1 $ \pm $ 2.2 &  78.3 $ \pm $ 1.2 &  \underline{49.1 $ \pm $ 2.5} & \underline{83.2 $ \pm $ 0.2} \\
\modelname\(_{8B-3I}\)  & \underline{65.1 $ \pm $ 2.4} & 78.3 $ \pm $ 0.4 & 59.5 $ \pm $ 2.1 & 79.3 $ \pm $ 0.4 & 51.6 $ \pm $ 2.2 & 79.6 $ \pm $ 0.5 & 48.2 $ \pm $ 1.7 & 82.9 $ \pm $ 0.3 \\ 
\modelname\(_{8B-3.1I}\) & \textbf{67.3 $ \pm $ 1.4} &  \underline{82.9 $ \pm $ 0.0} &  \textbf{64.7 $ \pm $ 2.1} &  \textbf{87.4 $ \pm $ 0.7} &  \underline{55.3 $ \pm $ 1.5} &  \textbf{84.3 $ \pm $ 0.5} &  \textbf{49.5 $ \pm $ 0.6} &  \textbf{84.5 $ \pm $ 0.1}  \\
\bottomrule
\end{tabular}}
\caption{Performance comparison of \modelname\ against baseline methods. The best-performing method is \textbf{bolded}, while the best baseline is \underline{underlined}. Results are reported as the average performance (\(\pm\) 1-std dev) over three runs, each using a different random seed, reported as a percentage (\%).}
\label{table:results}
\end{table*}

\subsection{Experimental Setup}
\label{subsec:experisetup}

We discuss the evaluation of \modelname\ and \discri, covering benchmark datasets, baselines, and evaluation metrics. Implementation details are discussed in Appendix \ref{app:implement}.

\paragraph{Benchmark Datasets.} We experiment on four open-source benchmarks: three from SemEval-2016 Task 13 Taxonomy Extraction Evaluation \cite{bordea-etal-2016-semeval}, Science (SemEval-Sci or Sci), Environment (SemEval-Env or Env) and Food (SemEval-Food) while the other is a collection of 114 WordNet sub-taxonomies (with 10 to 50 nodes each) \cite{bansal-etal-2014-structured} (refer to Table \ref{table:dataset} for statistics on benchmarks and Appendix \ref{app:benchmark} for train-test split and procedure to resolve definitions).

\paragraph{Baseline Methods.}We select two parallel baseline methods, one for \discri\ while another for \modelname. We compare the discriminative ranker, \discri, against commonly used ranking methods such as BM25, SimCSE and TEMP, Sentence-BERT, and BERT-Base-Uncased with cosine similarity. For \modelname's evaluation, we use 12 baselines, namely, (i) \textbf{BERT+MLP} \cite{devlin_bert_2019}, (ii) \textbf{TAXI} \cite{panchenko2016taxi}, (iii) \textbf{TaxoExpan} \cite{shen2020taxoexpan}, (iv) \textbf{STEAM} \cite{yu_steam_2020}, (v) \textbf{Musubu} \cite{takeoka-etal-2021-low} (vi) \textbf{TEMP} \cite{liu2021temp} (vii) \textbf{HEF} \cite{wang2021enquire} (viii) \textbf{BoxTaxo} \cite{jiang2023single}, (ix) \textbf{TaxoPrompt} \cite{xutaxoprompt}, (x) \textbf{TacoPrompt} \cite{xu-etal-2023-tacoprompt} (xi) \textbf{TaxoComplete} \cite{taxocomplete2023arous}, and (xii) FLAME \cite{flame2025}, which are discussed in Appendix \ref{app:baseline}.

\paragraph{Evaluation Metrics.} To evaluate the ranking performance of \discri, we use Hit@k as an evaluation metric, which is the number of correctly predicted parents in the \textit{top-k}. Since the ranked candidates are chunked, these metrics provide insight into whether the correct element appears within the \textit{top-k} positions. For evaluating \modelname, we follow the evaluation strategy used in CodeTaxo, which uses Accuracy and Wu \& Palmer similarity (Wu\&P) (discussed in Appendix \ref{app:eval}).

\begin{figure}[!t]
    \centering
    \includegraphics[width=1.0\linewidth]{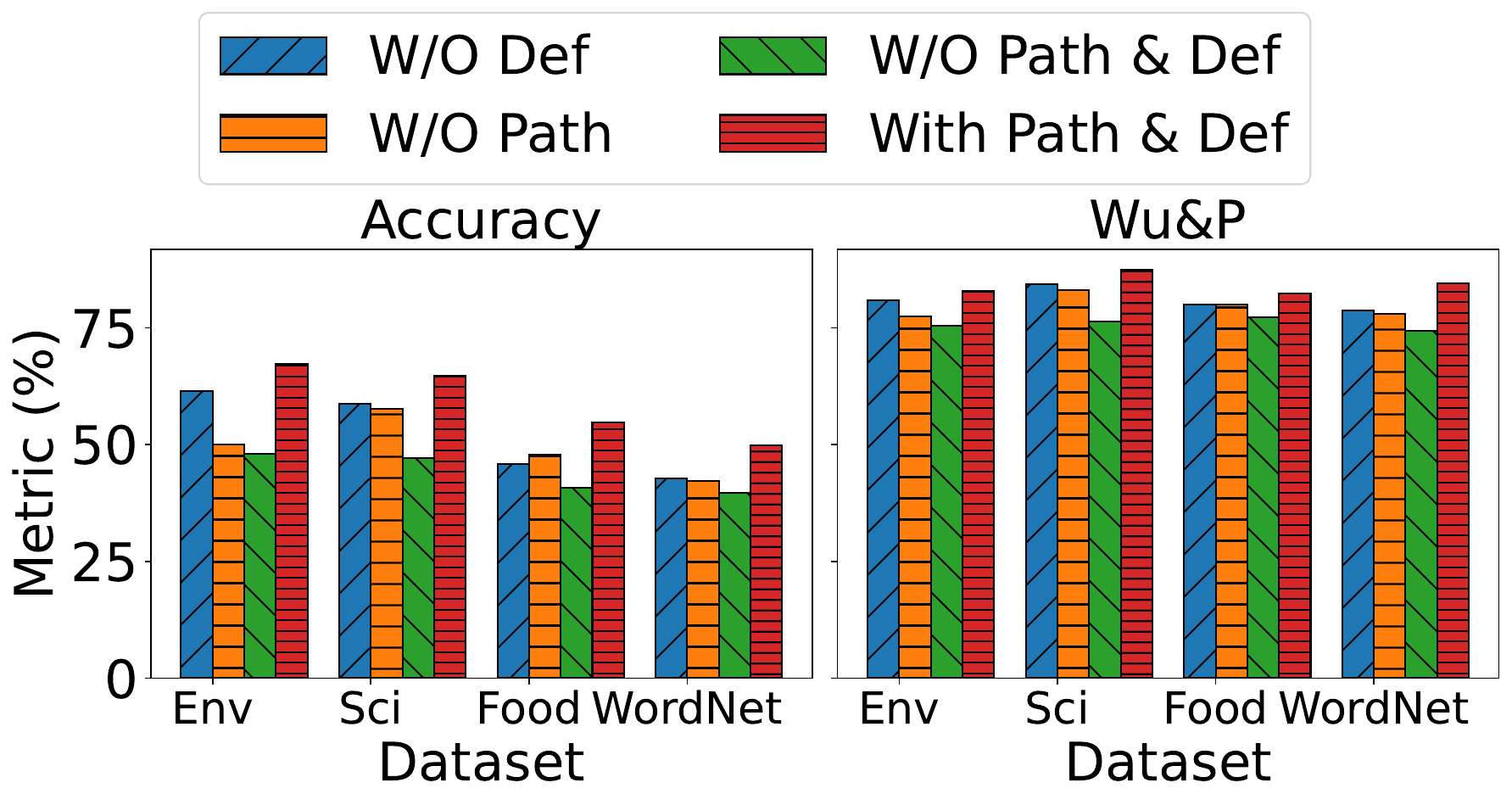}
        \caption{Performance comparison of hierarchical and semantic reasoning on parent retrieval across all benchmarks. Here W/O means \textbf{without}.}
    \label{fig:def_path}
\end{figure}

\begin{table}[!t]
\centering
\setlength{\tabcolsep}{3pt} 
\renewcommand{\arraystretch}{1.1}
\scalebox{0.78}{\begin{tabular}{ccccccc}
\toprule
\multirow{2}{*}{\textbf{Method}}       & \multicolumn{2}{c}{\textbf{Env}} & \multicolumn{2}{c}{\textbf{Sci}} & \multicolumn{2}{c}{\textbf{WordNet}} \\
                                       & \textbf{Acc}   & \textbf{Wu\&P}  & \textbf{Acc}   & \textbf{Wu\&P}  & \textbf{Acc}     & \textbf{Wu\&P}    \\ \midrule
\discri\ & 67.31          & 82.99           & 64.71          & 87.42           & 49.78            & 84.58             \\
Random Shuffle                         & 5.76           & 47.64           & 2.35           & 48.95           & 1.51             & 38.95             \\ \bottomrule
\end{tabular}
}
\caption{Effect of discriminative ranker on the performance of \modelname\ across ``Env'', ``Sci'' and ``WordNet'' benchmarks.}
\label{tab:discrirank}
\end{table}

\subsection{Performance Comparison}
\label{subsec:main}

\textbf{\discri.} As shown in Table \ref{table:results1}, our \discri\ model outperforms the best baseline, TEMP, across all Hit@k metrics on Env and Sci benchmarks (Refer to Table \ref{table:results2} in Appendix \ref{app:tempora} for results on other benchmarks). Specifically, it achieves a 23.5\% improvement in Hit@1 and a 19.1\% increase in Hit@10. From this, we can conclude that path-based rankers outperform semantic rankers. The enhanced ranking accuracy directly translates to improved efficiency and reduced latency in both the retrieval and verification stages.

\paragraph{\modelname.} Table \ref{table:results} shows the performance of \modelname\ on four benchmarks. As discussed in Appendices \ref{app:parprompt} and \ref{app:verprompt}, our framework uses prompts in an instruction-tuned paradigm. Therefore, we use three instruction-tuned LLMs , LLaMA-3.1-8B-Instruct (\modelname\(_{8B-3.1I}\)), LLaMA-3-8B-Instruct (\modelname\(_{8B-3I}\)), and LLaMA-2-7B-Chat (\modelname\(_{7B-2C}\)) to compare against state-of-the-art methods. As shown, \modelname\ surpasses all first and second-generation discriminative and third-generation prompting methods, with \modelname\(_{8B-3.1I}\) achieving the best performance. Specifically, \modelname\ achieves 12\% improvement in accuracy and 5\% improvement in Wu\&P over the best extensively trained generative baseline, FLAME.

\subsection{Ablation Studies}
\label{subsec:ablation}
We conduct ablations to analyze every component of \modelname. Specifically, we analyze (i) the impact of hierarchical and semantic reasoning on parent retrieval, (ii) effect of different types of reasoning verifiers, (iii) the influence of discriminative ranker, (iv) varying chunk sizes on the overall performance of \modelname, and (v) impact of semantic filter module on the performance and latency of \modelname. 

\begin{figure}[!t]
    \centering
    \includegraphics[width=1.0\linewidth]{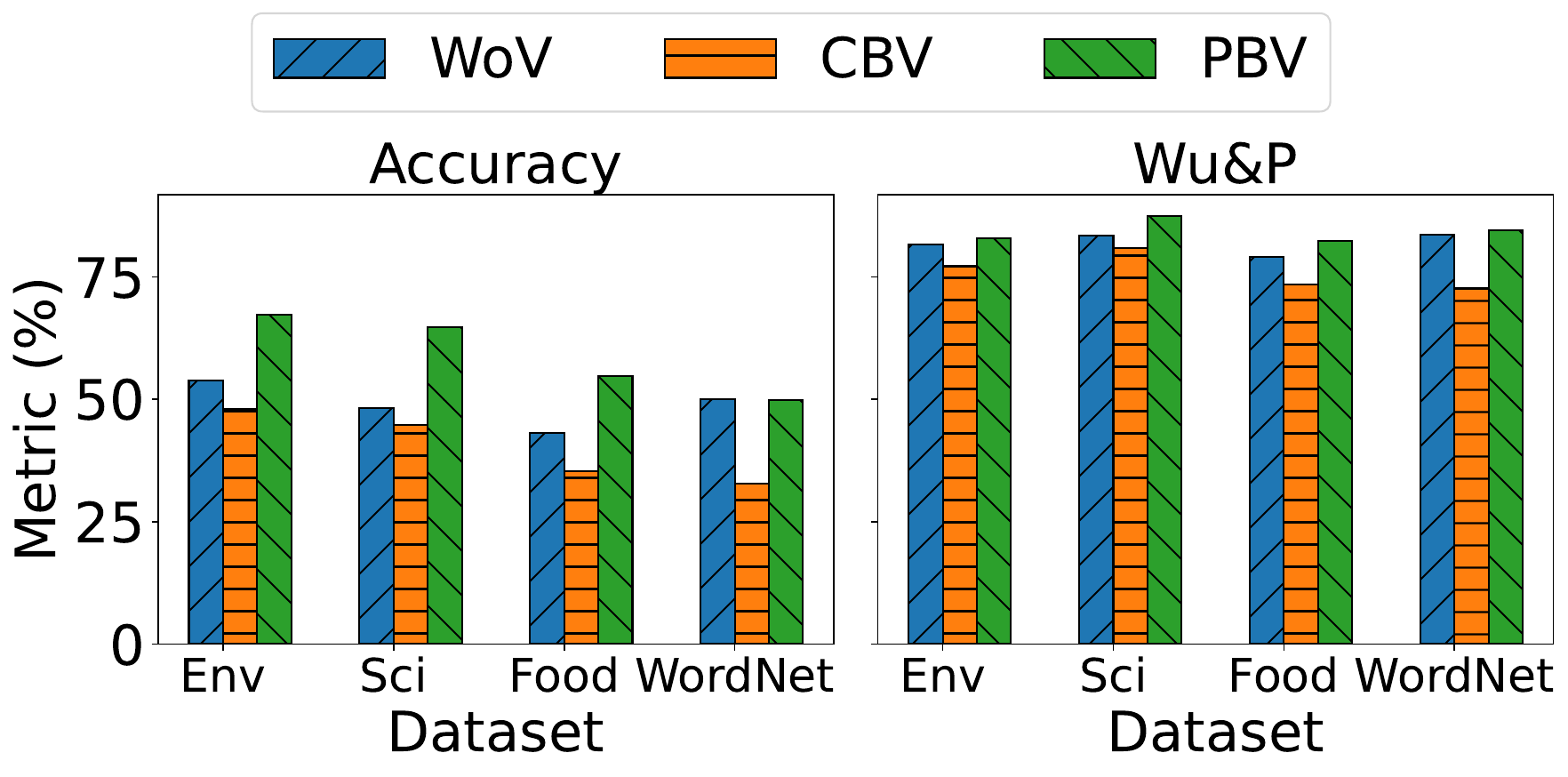}
    \caption{Performance comparison of different types of verifiers on SemEval-2016 benchmark. "WoV" = without verifier, "CBV" denotes candidate-based verifier, while "PBV" denotes path-based verifier.}
    \label{fig:verifier}
\end{figure}

\paragraph{Impact of Hierarchical and Semantic Reasoning on Parent Retrieval.} We analyze the impact of hierarchical and semantic reasoning by utilizing paths and definitions, on parent retrieval in \modelname, as shown in Fig. \ref{fig:def_path}. Using both paths and definitions gives the highest performance while removing both components leads to the worst performance, dropping accuracy by 28.3\% and Wu\&P by 5.3\%. Removing definitions reduces accuracy by 17.50\% and Wu\&P by 2.78\%. Further, removing hierarchical paths exacerbates the performance degradation, leading to a 31.99\% decrease in accuracy and a 9.78\% reduction in Wu\&P similarity scores.

\paragraph{Effect of Different Types of Reasoning Verifiers on \modelname's Performance.} Fig. \ref{fig:verifier} compares \modelname's performance under different verification strategies. The path-based verifier (PBV) achieves the highest accuracy, outperforming the candidate-based verifier (CBV) \footnote{CBV refers to a verifier setup where the LLM is directly prompted to rank candidate terms without access to their hierarchical paths. The ranking is based solely on average log token probabilities over candidate names, without structured reasoning over taxonomy paths.} by 47.04\% in accuracy and 10.84\% in Wu\&P scores. Even without a verifier (WoV), retrieval remains 21.31\% more accurate and 7.76\% higher in Wu\&P than CBV. The lowest performance is observed with CBV, where path information is absent, and verification relies solely on candidate evaluation, which introduces noise during verification and rejects relevant candidates. These results emphasize the importance of hierarchical reasoning during verification.

\begin{figure}[!t]
    \centering
    \includegraphics[width=1.0\linewidth]{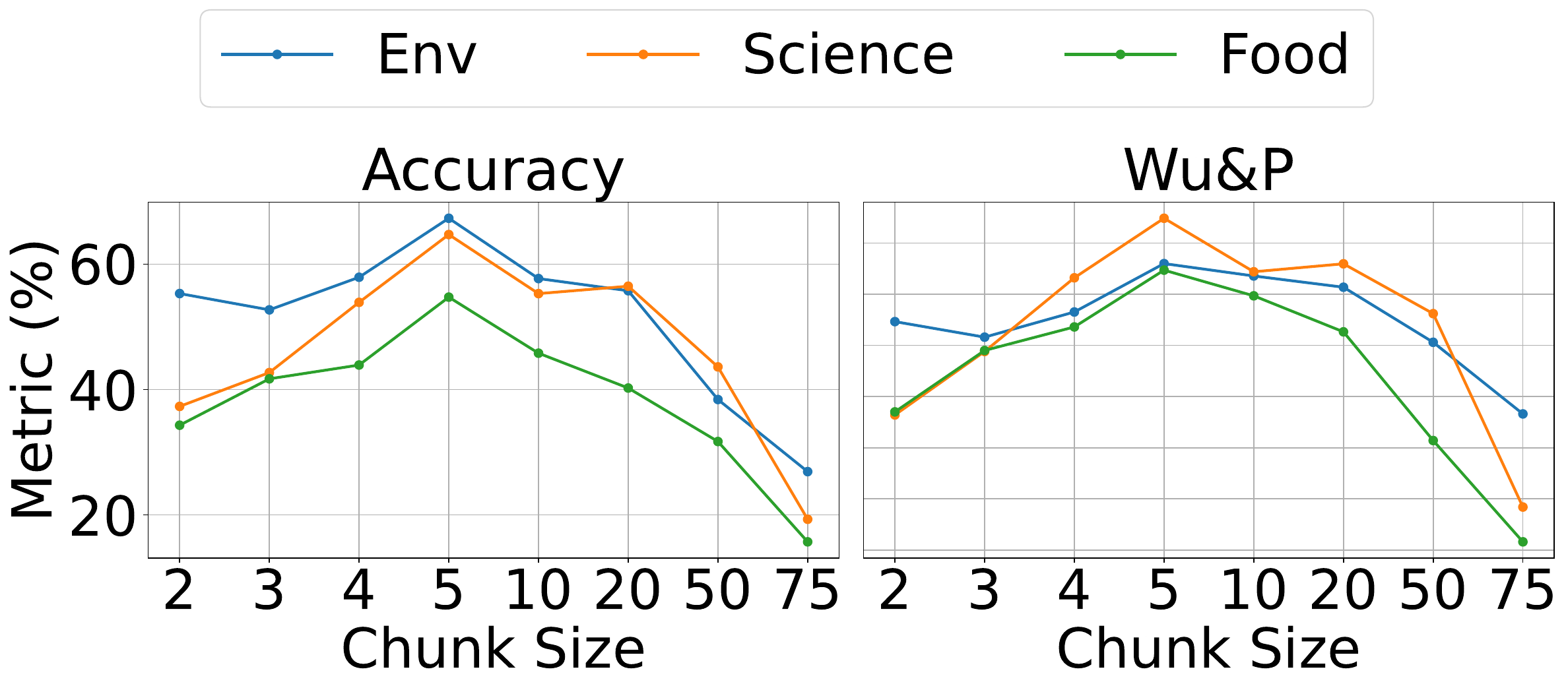}
    \caption{Effect of different chunk sizes on \modelname's performance across the SemEval-2016 benchmarks.}
    \label{fig:chunk}
\end{figure}

\paragraph{Impact of Discriminative Ranker \modelname's Performance.} Table \ref{tab:discrirank} compares \modelname's performance with and without the use of a discriminative ranker. The discriminative ranker improves accuracy by 57.33\% and Wu\&P scores by 39.8\%, highlighting its role in filtering relevant candidates and enhancing retrieval. In contrast, random shuffling of candidates disrupts the semantic ordering, often placing irrelevant or overly broad terms ahead of suitable hypernyms. This disordering hinders the retriever and verifier from progressing to later chunks where correct candidates may reside, thereby degrading overall system performance.

\begin{table}[!t]
\centering
\footnotesize
\setlength{\tabcolsep}{3pt} 
\renewcommand{\arraystretch}{1.1}
\begin{tabular}{lcc}
\toprule
\textbf{Setting} & \shortstack{\textbf{Avg. Runs}\\\textbf{per Query}} & \shortstack{\textbf{Queries Resolved}\\\textbf{in $\leq$3 Runs for "Env"}} \\
\midrule
Semantic Filter & 1.7 & 41 / 52 \\
W/O Semantic Filter & 3.1 & 18 / 52 \\
\bottomrule
\end{tabular}
\caption{Impact of the semantic filter module on reasoning efficiency. W/O denotes without.}
\label{tab:semantic_filter_runs}
\end{table}

\begin{table}[!t]
\centering
\footnotesize
\begin{tabular}{lcc}
\toprule
\textbf{Dataset} & \shortstack{\textbf{Batches with} \\ \textbf{True Parent Skipped}} & \textbf{Total Batches} \\
\midrule
Env & 3 & 42 \\
Sci & 8 & 69 \\
Food & 13 & 238 \\
WordNet & 21 & 370 \\
\bottomrule
\end{tabular}
\caption{Frequency of instances where the semantic filter skipped batches containing the true parent term.}
\label{tab:true_parent_skipped}
\end{table}

\renewcommand{\arraystretch}{1.0}
\begin{table*}[t]
\footnotesize
\centering
\begin{tabular}{p{0.22\linewidth} p{0.22\linewidth} p{0.22\linewidth} p{0.22\linewidth}}
\toprule
\textbf{Query} & \textbf{Parent Retriever (Ground Truth)} & \textbf{Path-Based Verifier} & \textbf{Candidate-Based Verifier}\\
\midrule
\textbf{Arctic Ocean} \newline \textit{The Arctic Ocean is the smallest and shallowest of the world's \colorbox{yellow}{oceans}, located around the North Pole and mostly covered by \colorbox{yellow}{sea} ice.} & 

\textbf{ocean (ocean)} 
\newline \textit{The most appropriate direct hypernym of 'Arctic Ocean' is \colorbox{yellow}{'ocean.'}} &

\textbf{Arctic Ocean > ocean > geophysical environment > environment} 
\newline \textit{The given path is the most appropriate path for the query node 'Arctic Ocean'.} &
\textbf{geophysical environment} 
\newline \textit{The query node 'Arctic Ocean' is a type of \colorbox{yellow}{geophysical environment}.}\\
\midrule
\textbf{exhaustion of resources} \newline \textit{a resource is a source or supply from which a benefit is produced and that has some utility. resource \colorbox{yellow}{exhaustion} refers to the depletion of these resources due to overuse or unsustainable practices.} &
\textbf{natural resources (natural resources)} \newline \textit{The most appropriate direct hypernym of 'exhaustion of resources' is \colorbox{yellow}{'natural resources.'}} &
\textbf{exhaustion of resources > natural resources > resource > possession} \newline \textit{The given path is 'exhaustion of resources -> natural resources -> resource -> possession', which appropriately contextualizes the query.} &
\textbf{resource} \newline \textit{The query node 'exhaustion of resources' is a type of \colorbox{yellow}{resource}.}
\\
\midrule
\textbf{pteridology} \newline \textit{the scientific study of ferns and related \colorbox{yellow}{plants} known as pteridophytes. this field focuses on their classification, morphology, \colorbox{yellow}{anatomy, ecology}, distribution, and evolutionary history.} & 

\textbf{biology (botany)} 
\newline \textit{The most appropriate direct hypernym of 'pteridology' is \colorbox{yellow}{'biology.'}} &

\textbf{pteridology > botany > biology > life science > natural science > science } 
\newline \textit{The query node 'pteridology' is the scientific study of ferns. The most direct and relevant hypernym of 'pteridology' is \colorbox{yellow}{'botany'}, which is the scientific study of plants.}
 &
\textbf{botany} 
\newline \textit{The query node 'pteridology' is the scientific study of ferns. The given candidate 'biology' is too broad as it encompasses the study of all living organisms. The alternative candidate \colorbox{yellow}{'botany'} is more specific.}\\
\midrule
\textbf{realpolitik} \newline \textit{realpolitik is a \colorbox{yellow}{political approach} that prioritizes practical considerations, and national interests over ideological concerns.} & 

\textbf{politics (politics)} 
\newline \textit{The most appropriate direct hypernym of 'realpolitik' is \colorbox{yellow}{'politics.'}} &

\textbf{realpolitik > politics > social science > science} 
\newline \textit{The given path is the most appropriate path for the query node 'realpolitik'.}
 &
\textbf{social science} 
\newline \textit{The most appropriate candidate for the query node 'realpolitik' is \colorbox{yellow}{'social science'.}}\\
\bottomrule
\end{tabular}
\caption{Case study analyzing the impact of \modelname's components on the "Env" and "Sci" benchmarks. The table presents component outputs in \textbf{bold}, with their definitions and explanations in \textit{italics}. Yellow highlighted text denotes key phrases in definitions and explanations that play an important role in reasoning.}
\label{table:casestudy}
\end{table*}

\paragraph{Effect of Different Chunk Sizes on \modelname's performance.} We analyze the impact of varying chunk sizes on \modelname's retrieval performance across SemEval-2016 benchmarks, as shown in Fig. \ref{fig:chunk}. The results indicate that smaller chunk sizes (2–3) lead to performance degradation due to limited contextual information, while excessively large chunk sizes ($\ge$20) reduce effectiveness as the retriever encounters an increased number of irrelevant candidates, introducing noise in the chunk. The best performance is achieved with chunk sizes between 4 and 10, indicating a balance between relevant context depth and noise.

\paragraph{Impact of Semantic Filter Module on Performance and Latency of \modelname.} As shown in Table~\ref{tab:semantic_filter_runs}, incorporating semantic filtering reduces the average number of retrieval-verification iterations per query from 3.1 to 1.7, and significantly increases the number of queries resolved within three runs. This demonstrates its utility in pruning irrelevant candidate batches early in the pipeline, reducing computational overhead and improving the latency of the framework. Furthermore, Table \ref{tab:true_parent_skipped} shows that the semantic filter rarely eliminates batches containing the true parent node, only 3 out of 42 batches (for batch size=5) in the Environment dataset and similarly low rates across other datasets, indicating that the filter maintains high recall while enhancing precision. Failure cases of semantic filtering are discussed in Appendix \ref{app:failsemantic}.

\subsection{Case Studies}
\label{subsec:case}
Table \ref{table:casestudy} discusses case studies for "Env" and "Sci" datasets, highlighting the reasoning mechanisms of \modelname's components. We select two samples of each dataset and observe that parent retriever mostly reasons over paths to directly select the most appropriate parent from candidate terms as visible in case of query terms ``\textit{Arctic Ocean}'' and ``\textit{exhaustion of resources}.'' However, it fails to retrieve the correct hypernym in case of ``\textit{pteridology}'' whose definition does not clearly define the term, creating confusion for parent retriever.  Moreover, the path-based verifier effectively exploits paths and reasons over them to distinguish between the retrieved path and alternatives, which is clearly evident in the case of ``\textit{pteridology},'' where the best-inferred path is ``\textit{pteridology > botany > biology > life science > natural science > science}.'' In contrast, the candidate-based verifier relies only on definitions, which are noisy, incomplete and incorrect, to verify the retrieved candidate, as seen in the case of ``\textit{realpolitik},'' where it fails to capture the appropriate hypernym, giving ``\textit{social science}'' as the hypernym. These results emphasize the importance of path-based reasoning in verification after candidate retrieval over verification using definitions, which introduces unnecessary abstraction. More such cases are discussed in Appendix \ref{app:case}.

\section{Conclusion}
We introduce \modelname, a plug-and-play taxonomy expansion framework which integrates discriminative ranking and generative reasoning to effectively retrieve and verify parent-child relationships without requiring fine-tuning of large language models. By leveraging the discriminative ranker \discri, the framework ranks candidate terms and optimally chunks them for LLM processing, ensuring relevant candidates are retained while maintaining context length constraints. Extensive experiments highlight that \modelname\ outperforms the best baseline. Ablation studies highlight the importance of hierarchical reasoning, discriminative ranking, and verification strategies, which improve retrieval quality. Case studies further validate \modelname's effectiveness in resolving ambiguous classifications and reinforcing hierarchical consistency.

\section*{Acknowledgements}
We gratefully acknowledge the support of Microsoft through the Microsoft Academic Partnership Grant 2024 (MAPG'24). T. Chakraborty acknowledges the support of the Rajiv
Khemani Young Faculty Chair Professorship in Artificial Intelligence.
\section*{Limitations}
While \modelname\ effectively retrieves parent terms in taxonomy expansion, its performance heavily relies on the quality of the candidate ranking. If the correct parent term is not highly ranked, \modelname\ suffers from increased retrieval errors and higher computational overhead due to multiple iterations of retrieval and verification. This iterative process significantly impacts both latency and accuracy, particularly in cases where the ranker fails to surface relevant candidates early. As a result, the overall system efficiency is constrained by the discriminative ranker, making \modelname\ highly sensitive to ranking quality. In other words, \textit{``the method is only as good as its ranker.''} Exploring better ranking techniques and reducing the reliance of retrievers on rankers will be the focus of our future works. We will also focus on improving interpretability by employing an ensemble of LLMs to reason over hierarchies. While this work focuses on taxonomy expansion, extending the framework to broader tasks such as taxonomy completion presents a promising direction for future research.

\section*{Ethical Considerations}
Our study uses open-source taxonomies like WordNet, which may include words that are sensitive, controversial, or offensive. These words fall into different offensive categories, such as mental health terms, violent actions, political topics, religious groups, law enforcement agencies, and negative or insulting words. Some terms related to social and economic status, extreme beliefs, or reputation damage may also carry unintended meanings or reinforce biases. Since large language models can sometimes generate biased or inappropriate hypernyms, we take steps to make taxonomy expansion more responsible. To prevent misleading or harmful word relationships, we limit the model's hypernym choices to a predefined set from the seed taxonomies. This ensures that the generated hierarchies remain accurate and unbiased.

\newpage
\bibliography{acl_latex}
\clearpage
\appendix
\section*{Appendix}
\section{Modified Euler Tour}
\label{app:hierarchy}

\begin{figure}[!b]
    \centering
\includegraphics[width=0.8\linewidth]{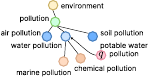}
    \caption{A hierarchical path of "water pollution" from "environment" taxonomy.}
    \label{fig:taxonomy}
\end{figure}

\begin{figure}[!b]
    \centering    \includegraphics[width=0.7\linewidth]{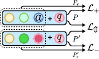}
    \caption{An illustration of loss functions defined in Section \ref{subsec:tempora}.}
    \label{fig:loss}
\end{figure}

As discussed in Section \ref{subsec:tempora}, we present an example of modified Euler-toured path for taxonomy shown in Fig. \ref{fig:taxonomy}. The Euler toured path is defined as \(P\) = ["water pollution", "pollution", "environment", "pollution", "air pollution", "pollution", "soil pollution", "pollution", "water pollution", "marine pollution", "water pollution", "chemical pollution"]. Therefore, the two types of verbalized paths (with and without query definition are discussed in Prompt 1 as shown in Fig. \ref{prompt:verb}, where \(P \in P_v(q,P)\) while \(P_r \in P_v(P)\). These paths are used to compute three types of losses as shown in Fig. \ref{fig:loss}.

\begin{figure*}[t]
\centering
\begin{mybox1}{Verbalized Paths}
\begin{mybox2}{\(P\)}
\sffamily
\small
potable water pollution is the contamination of water that is intended for human consumption [SEP] "water pollution" is child of "pollution" is child of "environment" is parent of "pollution" is parent of "air pollution" is child of "pollution" is parent of "soil pollution" is child of "pollution" is parent of "water pollution" is parent of "marine pollution" is child of "water pollution" is parent of "chemical pollution"
\end{mybox2}
\begin{mybox2}{\(P_r\)}
\sffamily
\small
"water pollution" is child of "pollution" is child of "environment" is parent of "pollution" is parent of "air pollution" is child of "pollution" is parent of "soil pollution" is child of "pollution" is parent of "water pollution" is parent of "marine pollution" is child of "water pollution" is parent of "chemical pollution"
\end{mybox2}
\end{mybox1}
\caption{Prompt 1 -- Verbalized Path Examples of $P$ and $P_r$.}
\phantomsection
\label{prompt:verb}
\end{figure*}

\begin{figure*}[t]
\centering
\begin{tcolorbox}
\sffamily
\small
You are a semantic relevance expert for terms present in a taxonomy. Your task is to determine whether the set of candidate terms is a semantically relevant match for the given query term 'Arctic Ocean' and 'Environment' taxonomy.\\

List of Candidate terms:\\
- geophysical environment\\
- ocean\\
- wild mammal\\
- animal life\\
- climatic zone\\

Definitions of candidate terms:\\
- geophysical environment - environment most often refers to:\\
- ocean - an ocean is a body of water that composes much of a planet's hydrosphere\\
- wild mammal - mammals, fur or hair, and three middle ear bones\\
- animal life - animals  are multicellular eukaryotic organisms that form the biological kingdom animalia\\
- climatic zone - climate classification systems are ways of classifying the world's climates\\

Reason over the definitions of candidate terms to determine their relevance as a whole with respect to query terms Arctic Ocean and Environment taxonomy.  If at least one of the candidate terms is the most relevant parent term for the taxonomy, return YES. Otherwise, return NO.\\

Answer: 
\end{tcolorbox}
\caption{Prompt 2 -- Semantic Filter Prompt Example}
\label{prompt:semfilexa}
\end{figure*}

\begin{figure*}[!t]
\centering
\begin{tcolorbox}
\sffamily
\small
You are an expert in hypernymy (is-a) relationship detection for a taxonomy. Your task is to find the most appropriate candidate hypernym of the query node 'Arctic Ocean' within the 'environment' taxonomy. The most appropriate hypernym is the most granular category that directly encompasses the query term.\\

List of Candidate terms:\\
- geophysical environment\\
- ocean\\
- wild mammal\\
- animal life\\
- climatic zone\\

Definitions of candidate terms:\\
- geophysical environment - environment most often refers to:\\
- ocean - an ocean is a body of water that composes much of a planet's hydrosphere\\
- wild mammal - mammals, fur or hair, and three middle ear bones\\
- animal life - animals  are multicellular eukaryotic organisms that form the biological kingdom animalia\\
- climatic zone - climate classification systems are ways of classifying the world's climates\\

Path from query term to root node for all candidate terms:\\
- Arctic Ocean -> geophysical environment -> environment\\
- Arctic Ocean -> ocean -> geophysical environment -> environment\\
- Arctic Ocean -> wild mammal -> animal life -> wildlife -> environment\\
- Arctic Ocean -> animal life -> wildlife -> environment\\
- Arctic Ocean -> climatic zone -> climate -> environment\\

Query Node Definition: The Arctic Ocean is the smallest and shallowest of the world's oceans.\\

Some examples of hypernymy relationships in the taxonomy are as follows:\\
- Children of geophysical environment are: arid zone, estuary, island, lake, mountain, ocean, plain, polar region, sea\\
- Parent of animal life is: wildlife\\
- Children of climatic zone are: equatorial zone, frigid zone, humid zone, subtropical zone, temperate zone, tropical zone\\
- Parent of climatic zone is: climate\\

Instructions:\\
- Determine the most appropriate direct hypernym of 'Arctic Ocean' from the given list of candidate terms. Do not return any term which is not in the list.\\
- A hypernym must be a most granular category in which the query node is an instance.\\
- If no candidate term correctly fits as a hypernym , return NOT FOUND.\\
- Do not include explanations, justifications, or additional context.\\

Answer:

\end{tcolorbox}
\caption{Prompt 3 -- Parent Retriever Prompt Example}
\label{prompt:paretrexa}
\end{figure*}

\begin{figure*}[!t]
\centering
\begin{tcolorbox}
\sffamily
\small
You are an expert verifier of hypernymy relationship for a taxonomy using paths. You have been given the following path from query term 'Arctic Ocean' to root node: Arctic Ocean -> ocean -> geophysical environment -> environment\\

Your task is to verify if the given path is the most appropriate path for the query node or the following paths provide a better alternative.\\

Other possible paths:\\
- Arctic Ocean -> geophysical environment -> environment\\
- Arctic Ocean -> ocean -> geophysical environment -> environment\\
- Arctic Ocean -> wild mammal -> animal life -> wildlife -> environment\\
- Arctic Ocean -> animal life -> wildlife -> environment\\
- Arctic Ocean -> climatic zone -> climate -> environment\\

Some examples of hypernymy relationships in the taxonomy are as follows:\\
- Children of geophysical environment are: arid zone, estuary, island, lake, mountain, ocean, plain, polar region, sea\\
- Parent of animal life is: wildlife\\
- Children of climatic zone are: equatorial zone, frigid zone, humid zone, subtropical zone, temperate zone, tropical zone\\
- Parent of climatic zone is: climate\\

Definitions of candidate terms:\\
- geophysical environment - environment most often refers to:\\
- ocean - an ocean is a body of water that composes much of a planet's hydrosphere\\
- wild mammal - mammals, fur or hair, and three middle ear bones\\
- animal life - animals  are multicellular eukaryotic organisms that form the biological kingdom animalia\\
- climatic zone - climate classification systems are ways of classifying the world's climates\\

Query Node Definition: The Arctic Ocean is the smallest and shallowest of the world's oceans.\\

Instructions:
- Compare the given path with the alternative paths based on definitions and hypernymy relationships.\\
- Select the most appropriate path that best represents the hierarchical relationship of 'Arctic Ocean' in the taxonomy.\\

Return the most appropriate path.\\

Answer:
\end{tcolorbox}
\caption{Prompt 4 -- Parent Verifier Prompt Example}
\label{prompt:paveriexa}
\end{figure*}

\section{Semantic Filter Prompt}
\label{app:semprompt}

The semantic filter module, discussed in Section \ref{subsec:semfilt}, relies on LLM to check if the chunk of candidates is relevant by reasoning over their definitions. The prompt example is discussed in Prompt 2 shown in Fig.~\ref{prompt:semfilexa}.

\section{Parent Retrieval Prompt}
\label{app:parprompt}

The parent retriever module, discussed in Section \ref{subsec:parentretriever}, reasons over taxonomy paths to determine the appropriate parent from the chunk of candidate terms. An example of the parent retriever prompt is discussed in Prompt 3 as shown in Fig. ~\ref{prompt:paretrexa}.

\section{Parent Verifier Prompt}
\label{app:verprompt}

The parent verifier module verifies the appropriateness of the retrieved candidate by reasoning on paths as discussed in Section \ref{subsec:parentverifier}. An example of the prompt is discussed in Prompt 4 as shown in Fig. \ref{prompt:paveriexa}.

\renewcommand{\arraystretch}{0.9}
\begin{table}[!t]
\centering
\footnotesize
\begin{tabular}{lcccc}
\toprule
\textbf{Hyperparameter} & \textbf{Sci} & \textbf{Food} & \textbf{Env} & \textbf{WordNet} \\
\midrule
Batch Size     & 32  & 32  & 64  & 32  \\
Learning Rate  & 2e-5  & 1e-5  & 2e-5  & 1e-5  \\
Max Padding    & 350  & 350  & 200  & 350  \\
$d$ & 0.2  & 0.2  & 0.2  & 0.2  \\
$\epsilon$ & 1e-8  & 1e-8  & 1e-8  & 1e-8  \\
Epochs         & 20  & 50  & 20  & 20  \\
\bottomrule
\end{tabular}
\caption{Hyperparameter settings for different datasets.}
\label{table:hyperparams}
\end{table}

\begin{algorithm}
\caption{\modelname}\label{algorithm:1}
\KwData{candidate terms $\mathcal{N}^o$, query term ${n^i_c}$, large language model $M$, discriminative ranker \discri}
\KwResult{candidate term $n_p$}
    \tcp{Return sorted batches of candidates}
    $\mathcal{B} \leftarrow \discri(n^i_c,\mathcal{N}^o)$\;
    \tcp{Iterate through candidate batches}
    \For{$B^i_j \in \mathcal{B}$}{
        \If{$\text{Semantic-Filter}(B^i_j)$}{
            \While{$|B^i_j| > 1$}{
            $\hat{n}_p \leftarrow\textit{Parent-Retriever}(q,B^i_j)$\;
                \tcp{Perform path verification and update batch}
                $P^*_j \leftarrow\textit{Parent-Verifier}(q,B^i_j)$\;
                \If{$\hat{n}_p \in P^*_j$}{
                    \Return{$\hat{n}_p$};
                }
                \Else{
                    \tcp{Remove retrieved candidate and continue}
                    $B^i_j \gets B^i_j \setminus \{ \hat{n}_p \}$\;
                }
            }
        }
        $\mathcal{B} \gets \mathcal{B} \setminus B^i_j$\;
    }
    \Return{``Not Found''}
\end{algorithm}

\section{Pseudocode}
\label{app:pseudocode}

The pseudocode of \modelname\ framework, which is discussed in Section \ref{method}, is shown in Algorithm \ref{algorithm:1}. For a query term \(n^i_c\), the discriminator \discri\ first ranks and chunks all candidate terms \(\mathcal{N}^o\) as \(\mathcal{B}\). Then, we select a chunk \(B\) and apply the parent retriever to get the most appropriate parent \(\hat{n}_p\), followed by verification. If verification fails, the candidate is discarded, and the retrieval process is re-iterated on the remaining chunk. The iterative refinement continues until only one candidate remains in the chunk.

\section{Implementation Details}
\label{app:implement}
\begin{figure}[ht]
    \centering
    \includegraphics[width=0.9\linewidth]{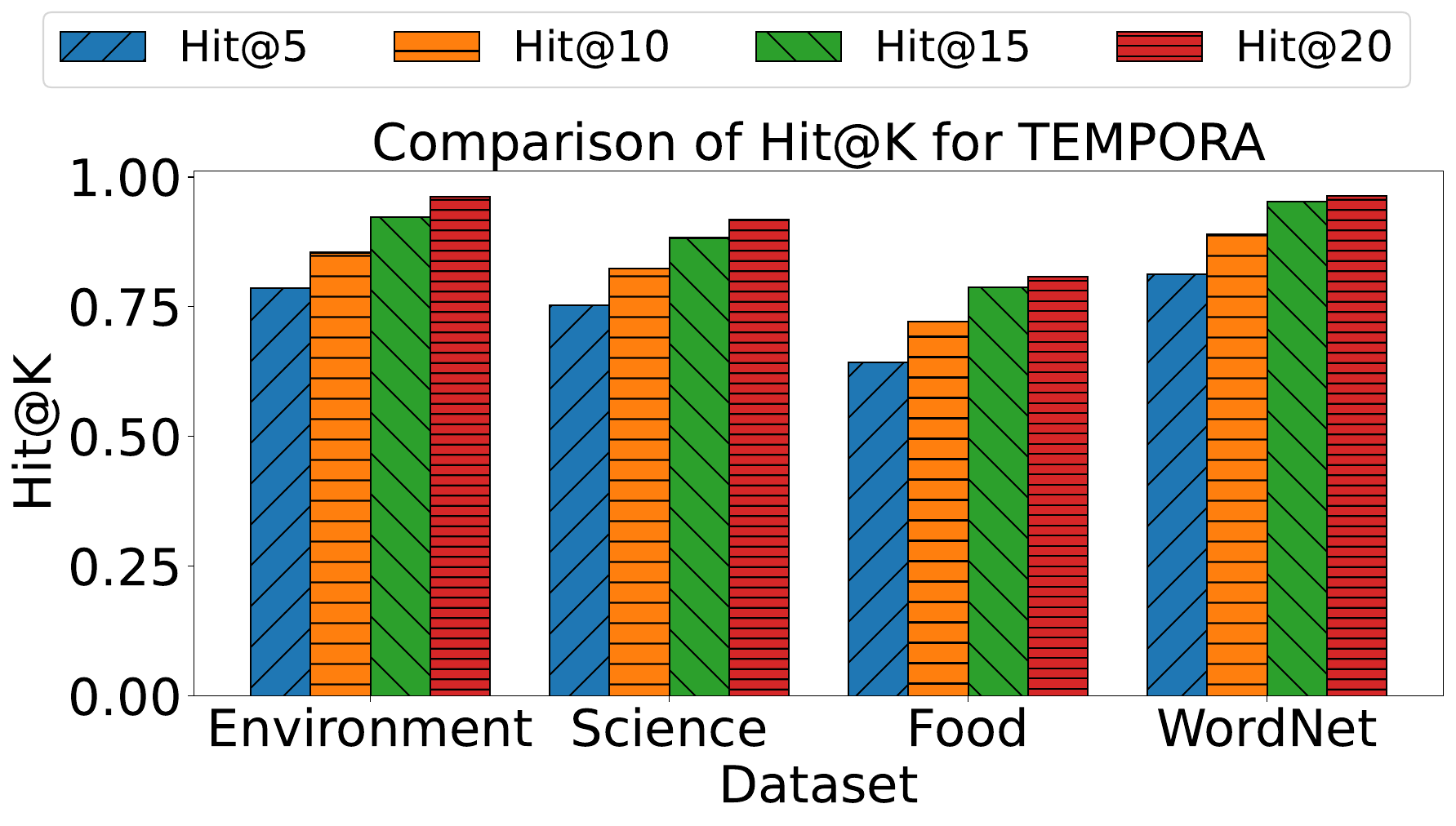}
    \caption{Comparison of Hit@K performance for \discri\ across different benchmarks (Environment, Science, Food, and WordNet).}
    \label{fig:hitatk}
\end{figure}

\modelname\  and BERT+MLP are implemented in PyTorch, with all other baselines obtained from their respective official repositories. The training and inference processes are executed on a single 80GB NVIDIA A100 GPU to ensure computational efficiency. To implement \discri, we use \texttt{bert-base-uncased} as the default pre-trained model, optimized using AdamW. The hyperparameters utilized for training and inference are detailed in Table \ref{table:hyperparams}. Notably, \discri\ is trained on all benchmarks for fewer epochs compared to TEMP, as the primary objective is to achieve a ranker that delivers sufficiently high performance without excessive fine-tuning. Retrieval and verification rely on instruction-tuned models from the LLaMA family, namely LLaMA-3.1-8B-Instruct, LLaMA-3-8B-Instruct, and LLaMA-2-7B-Chat. As demonstrated in Fig. \ref{fig:hitatk}, all benchmarks except Food achieve a Hit@15 score above 0.9, indicating that for 90\% of queries the correct parent exists within the top-15 candidates. Therefore, to enhance the efficiency of retrieval and verification, inference is performed on three chunks or 15 candidates simultaneously. If any chunk is empty, it is discarded, and only the remaining chunks are considered. This strategy ensures that the correct output is obtained in a single inference pass, optimizing both computational efficiency and retrieval accuracy. For text generation tasks, a stopping criterion is enforced to terminate generation upon encountering a newline character, which prevents the generation of unnecessary tokens, thereby reducing inference latency.

\renewcommand{\arraystretch}{0.9}
\begin{table}[!t]
\centering
\scalebox{0.9}{\begin{tabular}{lrrrr}
\toprule
Dataset           & Env & Sci & Food & \textcolor{black}{WordNet} \\ \midrule
$|\mathcal{N}^0|$ & 261 & 429 & 1486 & \textcolor{black}{20.5}    \\
$|\mathcal{E}^0|$ & 261 & 452 & 1576 & \textcolor{black}{19.5}    \\
$|D|$             & 6   & 8   & 8    & \textcolor{black}{3}       \\ \bottomrule
\end{tabular}}
\caption{Statistics of the four benchmark datasets. Here, $|\mathcal{N}^0|$ and $|\mathcal{E}^0|$ indicate the number of nodes and edges in the initial taxonomy, respectively, while $|D|$ represents the taxonomy depth. For WordNet, $|\mathcal{N}^0|$ and $|\mathcal{E}^0|$ denote the average number of nodes and edges across 114 sub-taxonomies.}
\label{table:dataset}
\end{table}

\renewcommand{\arraystretch}{0.9}
\begin{table}[!t]
\centering
\footnotesize
    \begin{tabular}{lcccc}
    \toprule
\textbf{Method} & \multicolumn{2}{c}{\textbf{Food}} & \multicolumn{2}{c}{\textbf{WordNet}} \\
\textbf{}       & \textbf{Hit@1}     & \textbf{Hit@10}     & \textbf{Hit@1}   & \textbf{Hit@10}   \\
\midrule
BM25            & 0.071              & 0.165               & 0.065            & 0.171             \\
SBERT$_{COS}$   & 0.182              & 0.380               & 0.113            & 0.390             \\
BERT$_{COS}$    & 0.077              & 0.118               & 0.076            & 0.225             \\
SimCSE          & 0.119              & 0.280               & 0.258            & 0.621             \\
TEMP            & \underline{0.330}  & \underline{0.529}   & \underline{0.329} & \underline{0.682} \\ 
\midrule
\discri\        & \textbf{0.451}     & \textbf{0.663}      & \textbf{0.405}   & \textbf{0.794}    \\ 
\bottomrule
\end{tabular}
\caption{Comparison of \discri\ with the discriminative ranking baseline methods. The best performance is marked in bold, while the best baseline is underlined.}
\label{table:results2}
\end{table}

\renewcommand{\arraystretch}{0.85}
\begin{table*}[!t]
\centering
\begin{tabular}{p{0.15\linewidth} p{0.22\linewidth} p{0.55\linewidth}}
\toprule
\textbf{Query} & \textbf{True Parent} & \textbf{Batch Skipped} \\
\midrule
fire protection & environmental protection & [\textbf{environmental protection}, environment, nuisance, environmental policy, pollution control measures] \\
desert & geophysical environment & [environment, degradation of the environment, \textbf{geophysical environment}, physical environment, wildlife] \\
demography & sociology & [social science, \textbf{sociology}, anthropology, economics, correlation] \\
thanatology & science & [endocrinology, anesthesiology, \textbf{science}, internal medicine, podiatry] \\
eugenics & genetics & [biology, \textbf{genetics}, life science, medical science, molecular biology]\\
scrumpy & cider & [\textbf{cider}, hard cider, sweet cider, alcohol, hooch]\\
\bottomrule
\end{tabular}
\caption{Failure cases where the LLM-based semantic filter incorrectly filters out batches with true parent.}
\label{tab:case_study_failures}
\end{table*}

\section{Benchmark Datasets}
\label{app:benchmark}
The statistics of benchmark datasets, as discussed in Section \ref{subsec:experisetup}, namely Environment, Science, Food and WordNet are shown in Table \ref{table:dataset}. For WordNet, the total number of nodes and edges are 2309 and 2226, respectively across 114 sub-taxonomies.  We use GPT-4o to correct missing, mislabeled, or corrupted definitions. Following \cite{liu2021temp}, we randomly select 20\% of leaf concepts for testing.

\section{Baseline Methods} 
\label{app:baseline}
As mentioned in Section \ref{subsec:experisetup}, we compare the performance of \modelname\ against twelve baseline methods.
\begin{itemize}[noitemsep, nolistsep, topsep=0pt, leftmargin=1em]
  \item \textbf{BERT+MLP} \cite{devlin_bert_2019} utilizes BERT to get term embeddings to predict hypernymy relationships.
  \item \textbf{TAXI} \cite{panchenko2016taxi}, a winner of SemEval-2016 Task 13 is a hypernym detection-based method that constructs taxonomies by identifying hypernymy relations between entity pairs based on lexical pattern extraction and substring matching.
  \item \textbf{TaxoExpan} \cite{shen2020taxoexpan} uses position-enhanced Graph Neural Networks (GNNs) to encode local structure. It models anchor representations using ego-network encoding and scores hypernymy relations with a log-bilinear feed-forward model, optimizing learning via the InfoNCE loss.
  \item \textbf{STEAM} \cite{yu_steam_2020} leverages mini-paths to model the local structure of the anchor nodes. It integrates graph-based, contextual, and manually crafted lexical-syntactic features for query-anchor pairs and employs multi-view co-training to enhance hypernymy detection.
  \item \textbf{Musubu} \cite{takeoka-etal-2021-low} leverages pre-trained models, fine-tuning them to classify hypernymy. It utilizes queries derived from Hearst patterns to guide the fine-tuning process.
  \item \textbf{TEMP} \cite{liu2021temp} leverages a pre-trained model to encode definitions of taxonomy concepts while integrating structural information through taxonomic paths optimizing through dynamic margin loss.
  \item \textbf{HEF} \cite{wang2021enquire} concatenates neighbor embeddings to construct the ego-tree representation and learns Stopper and Pathfinder scorers to identify candidate nodes in the taxonomy.
  \item \textbf{BoxTaxo} \cite{jiang2023single} employs box embeddings to represent hierarchical relationships and optimizes geometric and probabilistic losses to score parent nodes. It leverages the volume of hyper-rectangles to model taxonomic structures effectively.
  \item \textbf{TaxoPrompt} \cite{xutaxoprompt} prompts encoder models such as BERT to capture taxonomic structures. It formulates prompt templates using random walks to represent taxonomy context and optimizes the model via the masked language modeling objective.
  \item \textbf{TacoPrompt} \cite{xu-etal-2023-tacoprompt} employs multi-task learning to mitigate overfitting taxonomy completion tasks. It leverages prompt-based semantic triplet matching to enhance the model's generalization capabilities.
  \item \textbf{TaxoComplete} \cite{taxocomplete2023arous} integrates a query-anchor semantic matching mechanism with a direction-aware propagation module, facilitating a more accurate hierarchical representation of nodes.
  \item \textbf{FLAME} \cite{flame2025} fine-tunes low-rank parameters of LLMs to predict parent terms by leveraging semantically relevant local and global neighborhood information of candidates.
\end{itemize}

\renewcommand{\arraystretch}{0.85}
\begin{table*}[!ht]
\centering
\footnotesize
\begin{tabular}{p{0.22\linewidth} p{0.22\linewidth} p{0.25\linewidth} p{0.22\linewidth}}
\toprule
\textbf{Query} & \textbf{Parent Retriever (Ground Truth)} & \textbf{Path-Based Verifier} & \textbf{Candidate-Based Verifier}\\
\midrule
\textbf{grapefruit peel} \newline \textit{grapefruit peel has a bright, \colorbox{yellow}{citrusy} flavor with a balance of bitterness and sweetness.} & 
\textbf{candied citrus peel (candied citrus peel)} \newline \textit{The most appropriate direct hypernym of 'grapefruit peel' is \colorbox{yellow}{'candied citrus peel.'}} &
\textbf{grapefruit peel > candied citrus peel > candied fruit > confiture > sweet > dainty > nutriment > food} \newline \textit{The given path is the most appropriate path because it directly connects \colorbox{yellow}{'grapefruit peel' to the root node} \colorbox{yellow}{ 'food'} through a series of hypernymy relationships that are relevant to the definition of 'grapefruit peel'.} &
\textbf{lemon peel} \newline \textit{The definition of \colorbox{yellow}{lemon peel} is more specific and closely related to the query node 'grapefruit peel' compared to the other candidates.}\\
\midrule
\textbf{demography} \newline \textit{the scientific study of human populations, including their size, composition, distribution, density, and growth.} & 
\textbf{sociology (sociology)} \newline \textit{The most appropriate direct hypernym of 'demography' is \colorbox{yellow}{'sociology.'}} &
\textbf{demography > sociology > social science > science} \newline \textit{The path is appropriate because 'demography' is a subfield of \colorbox{yellow}{sociology}, which belongs to \colorbox{yellow}{social science}, a domain within \colorbox{yellow}{science}. This path reflects a logical hierarchical classification of 'demography'.} &
\textbf{social science} \newline \textit{Although \colorbox{yellow}{social science} is a broader category that includes demography, the term 'sociology' is more specific and directly connected to the study of population structures. Therefore, 'sociology' better captures the essence of the query than 'social science'.}\\
\midrule
\textbf{Irish soda bread} \newline \textit{A variety of \colorbox{yellow}{quick bread} traditionally made in a variety of cuisines in which sodium bicarbonate is used as a leavening agent instead of the traditional yeast.} & 
\textbf{bread (quick bread)} \newline \textit{The most appropriate direct hypernym of 'Irish soda bread' is \colorbox{yellow}{'bread.'}} &
\textbf{Irish soda bread > quick bread > bread > starches > foodstuff > food} \newline \textit{The query node 'irish soda bread' is a type of \colorbox{yellow}{'quick} \colorbox{yellow}{bread'} as per the definition. \colorbox{yellow}{'Quick bread' is a type of 'bread'} as per the hypernymy relationship. 'Bread' is a type of 'starches' as per the hypernymy relationship. 'Starches' is a type of 'foodstuff' as per the hypernymy relationship. 'Foodstuff' is a type of 'food' as per the hypernymy relationship.} &
\textbf{quick bread} \newline \textit{The query node 'irish soda bread' is a type of \colorbox{yellow}{quick bread} as it uses sodium bicarbonate as a leavening agent instead of traditional yeast.}\\
\midrule
\textbf{differential calculus} \newline
\textit{a branch of \colorbox{yellow}{calculus} concerned with the study of the rates at which quantities change.} & 
\textbf{calculus (calculus)} \newline
\textit{The most appropriate direct hypernym of 'differential calculus' is \colorbox{yellow}{'calculus'}.} & 
\textbf{differential calculus > calculus > pure mathematics > mathematics > science} \newline
\textit{The path traverses through \colorbox{yellow}{calculus} and correctly situates the concept in the broader scientific hierarchy.} & 
\textbf{calculus} \newline
\textit{The candidate \colorbox{yellow}{calculus} is appropriate.} \\
\midrule
\textbf{destruction of crops} \newline
\textit{a crop is a plant or animal product that can be grown and harvested for profit or subsistence. \colorbox{yellow}{Destruction} refers to damage to this resource.} & 
\textbf{degradation of environment (degradation of environment)} \newline
\textit{The most appropriate direct hypernym of 'destruction of crops' is \colorbox{yellow}{'degradation'} of environment.} & 
\textbf{destruction of crops > degradation of the environment > environment} \newline
\textit{This path is appropriate because \colorbox{yellow}{degradation of the environment} includes anthropogenic harm to agricultural systems.} & 
\textbf{degradation of environment} \newline
\textit{The candidate node \colorbox{yellow}{degradation of environment} is valid.} \\
\bottomrule
\end{tabular}
\caption{Case study analyzing the impact of \modelname's components on "Env", "Food" and "Science" benchmarks.}
\label{table:casestudy2}
\end{table*}

We adapt taxonomy completion baselines, such as TaxoComplete  and TacoPrompt for the taxonomy expansion setting by introducing a pseudo-placeholder in place of the child node. This transformation ensures that the task is framed as a leaf node insertion, allowing these completion methods, originally designed for intermediate node prediction, to be effectively applied to taxonomy expansion tasks.

\section{Evaluation Metrics}
\label{app:eval}

During inference, both the baselines and \modelname\ rank all candidate terms for each query node. Given a query set $\mathcal{C}$, the predictions are denoted as \(\left\{\hat{y}_1, \hat{y}_2, \cdots, \hat{y}_{|\mathcal{C}|}\right\}\) while the corresponding ground truths are represented as \(\left\{y_1, y_2, \cdots, y_{|\mathcal{C}|}\right\}\). Following CodeTaxo (\cite{zeng2024codetaxo}), we evaluate performance using two key metrics, as outlined in Section \ref{subsec:experisetup},
\begin{itemize}[noitemsep, nolistsep, topsep=0pt, leftmargin=1em]
    \item \textbf{Accuracy (Acc):} This metric quantifies the proportion of predicted candidates that correctly match the ground-truth, 
    \begin{equation}
      \text{Acc}=\frac{1}{|\mathcal{C}|}\sum_{i=1}^{|\mathcal{C}|}{\mathbb{I}\left(y_i=\hat{y_i}\right)},  
    \end{equation}
    where \(\mathbb{I}(\cdot)\) denotes the indicator function.
    
    \item \textbf{Wu \& Palmer Similarity (Wu\&P):} This metric quantifies the structural similarity between the predicted parent and the ground truth within the seed taxonomy. It is computed based on the closest common ancestor,
    \begin{equation}
        \text{Wu\&P}=\frac{1}{|\mathcal{C}|}\sum_{i=1}^{|\mathcal{C}|}{\frac{2\times \text{depth}\left(\text{LCA}\left(\hat{a_i}, a_i\right)\right)}{\text{depth}\left(\hat{a_i}\right)+\text{depth}\left(a_i\right)}},
    \end{equation}
    where $\text{depth}(\cdot)$ represents the depth of the node, and $\text{LCA}(\cdot,\cdot)$ denotes the least common ancestor of the predicted candidate and ground truth.
\end{itemize}

\section{Discriminator Results}
\label{app:tempora}

Table \ref{table:results2} provides additional empirical evidence supporting the performance of \discri, as analyzed in Section \ref{subsec:main}. The results indicate that \discri\ consistently outperforms TEMP across both evaluation metrics on the Food and WordNet benchmarks.

\section{Failure Cases of Semantic Filtering}
\label{app:failsemantic}

While the semantic filter module is effective in improving efficiency and recall (as discussed in Section~\ref{subsec:ablation}), there are a few failure cases where it incorrectly filters out batches containing the true parent term. These errors typically stem from ambiguity or insufficient granularity in the definitions used for semantic matching. As illustrated in Table~\ref{tab:case_study_failures}, for the query term \textit{fire protection}, the filter erroneously rejects the batch containing the correct parent \textit{environmental protection}, with the LLM reasoning that fire protection does not semantically fall under that category. Similar misclassifications are observed for terms like \textit{desert} and \textit{thanatology}, where broader parent categories such as \textit{geophysical environment} and \textit{science} were excluded due to narrow or domain-specific interpretations. These cases highlight the limitations of LLM-based semantic inference when reasoning over abstract or generalized definitions, though such instances remain infrequent across datasets.

\section{Case Studies}
\label{app:case}
Extending on the case studies discussed in Section \ref{subsec:case}, we further analyze some additional examples with more semantically rich explanations in Table \ref{table:casestudy2}. The parent retriever provides straightforward reasoning over definitions and path-based relationships. However, as it is susceptible to errors, we rely on the verifier's reasoning to validate and refine the retriever's predictions. The path-based verifier employs lineage-oriented reasoning to assess retrieval accuracy by analyzing hierarchical paths. For instance, in the case of `grapefruit peel', the verifier explains, ``The given path is the most appropriate as it directly connects `grapefruit peel' to the root node `food' through a series of hypernymy relationships that align with its definition.'' In contrast, the candidate-based verifier relies on definitions, which can lead to incorrect predictions when contextual hierarchical relationships are not explicitly captured in definitions.

\end{document}